%% file: acl2023.tex
\pdfoutput=1

\documentclass[11pt]{article}

\usepackage{ACL2023}

\usepackage{times}
\usepackage{latexsym}

\usepackage[T1]{fontenc}

\usepackage[utf8]{inputenc}

\usepackage{microtype}
\usepackage{graphicx}

\usepackage{tikz}

\usepackage{inconsolata}

\title{Understanding BLOOM: An empirical study on diverse NLP tasks}

\author{Parag Pravin Dakle, SaiKrishna Rallabandi, Preethi Raghavan\\
  AI Center of Excellence, Fidelity Investments, Boston MA\\
  \{\texttt{paragpravin.dakle, saikrishna.rallabandi, preethi.raghavan\}@fmr.com}}

\begin{document}
\maketitle
\begin{abstract}

We view the landscape of large language models (LLMs) through the lens of the recently released BLOOM model to understand the performance of BLOOM and other decoder-only LLMs compared to BERT-style encoder-only models. We achieve this by evaluating the smaller BLOOM model variants (\textit{350m/560m} and \textit{1b3/1b7}) on several NLP benchmark datasets and popular leaderboards. We make the following observations: (1) BLOOM performance does not scale with parameter size, unlike other LLMs like GPT and BERT. Experiments fine-tuning BLOOM models show that the 560m variant performs similarly to or better than the 1b7 variant, (2) Zero-shot cross-lingual and multi-lingual fine-tuning experiments show that BLOOM is at par or worse than monolingual GPT-2 models, and (3) Toxicity analysis of prompt-based text generation using the RealToxicityPrompts dataset shows that the text generated by BLOOM is at least 17\% less toxic than GPT-2 and GPT-3 models.

\end{abstract}

\section{Introduction}
\input{sections/introduction}

\section{Related Work}
\input{sections/related_works}

\section{Motivation}
\label{sec:motivation}

Our primary goal is to evaluate and benchmark the recently released family of BLOOM models on a broad set of fundamental NLP tasks and compare its performance with other popular LLMs. We choose tasks that test the model(s) both syntactically and semantically, focusing more on the latter. 

We carry out a syntactic evaluation by checking grammatical correctness, and for semantic evaluation, we consider the following tasks:

\noindent
\textbf{Paraphrase Detection (PD). }
This consists of a binary paraphrase detection task and a level of semantic similarity detection on a scale of 1-5.

\noindent
\textbf{Textual entailment (TE).} Given two text fragments, this task determines if the meaning of one text is entailed from another text. We consider entailment in various forms: (i) Direct - Binary (entailment/not entailment) and ternary (entailment/contradiction/neutral), (ii) Question Answering - In sentence-question pair, determine if the sentence contains the answer to the question, and (iii) Coreference Resolution - Given a sentence pair where the second sentence contains a pronominal coreference resolution, is the second sentence entailed by the first?

\noindent
\textbf{Question Answering.} We test a model's ability to do question answering using four tasks: 1)  Question classification to understand how well a model can identify question types, 2) Answer detection to test if the model can detect if a given sentence contains an answer to a given question, 3) Question answering to see if a model can answer a question from a given context, and 4) Answer generation to evaluate a model's answer generation capability and if it can remember facts seen during training.

\noindent
\textbf{Topic Understanding.} We test to see if a model can understand the main topic or theme of the sentence. We also use the task of article generation given the title towards this evaluation.

\noindent
\textbf{Entity Extraction.} The task of NER is used to evaluate a model's ability to identify spans that share similar semantic characteristics.

In addition to evaluating a model's syntactic and semantic understanding, this study also explores -

\noindent
\textbf{Multilingual tasks.} Since BLOOM models are trained on texts from 46 languages, we do cross-language evaluation by training on English and testing on German, and multilingual evaluation by training and testing using a mixture of 6 languages.

\noindent
\textbf{Zero- and Few-shot learning.} We evaluate the BLOOM model variants on prompt-based zero- and few-shot learning. The work of \citet{scao2022bloom} extensively focuses on evaluating the BLOOM models on various datasets using zero- and one-shot learning. However, the authors only focus on zero- and one-shot compared to this work, which includes four- and eight-shot. Additionally, \citeauthor{scao2022bloom} do not evaluate the BLOOM model(s) on text extraction tasks in zero- and few-shot settings. The work presented in this paper includes text extraction tasks along with text classification tasks.

\noindent
\textbf{Toxicity evaluation.} We evaluate BLOOM  with regard to generating toxic language using prompts.   

\section{Experiments and Results}\label{sec:experiment_results}
This section focuses on experiments and results evaluating the BLOOM variants on previously identified tasks. Table \ref{tab:dataset_list} lists the datasets used for the evaluation and provides their mapping to the motivation tasks. The baseline models used for evaluation are described below:

\textbf{BERT-style} - We consider BERT \cite{devlin-etal-2019-bert}, RoBERTa \cite{DBLP:journals/corr/abs-1907-11692}, DistilBERT \cite{Sanh2019DistilBERTAD}, and DeBERTa V3 \cite{he2021debertav3} models. For experiments with English, we use \textit{bert-base-uncased}, \textit{roberta-base}, \textit{distilbert-base-uncased}, and \textit{deberta-v3-base} models as the baselines. All multi-lingual experiments use \textit{bert-base-multilingual-cased}, \textit{xlm-roberta-base} \cite{conneau-etal-2020-unsupervised}, and \textit{distilbert-base-multilingual-cased} models.

\textbf{GPT-style}: Three GPT-2 \cite{radford2019language} variants have been used as baselines - \textit{gpt2} or \textit{gpt2-small}, \textit{gpt2-medium}, and \textit{gpt2-xl}. GPT-Neo \cite{gpt-neo} is a Transformer based model designed using EleutherAI's replication of the GPT-3 architecture and the \textit{gpt-neo-1.3B} variant is used for text generation and few-shot learning experiments.
    
\textbf{BLOOM}: For all experiments except text generation, the smallest two variants of the BLOOM model are used as baselines. When the BLOOM model was launched, the two smallest variants were \textit{bloom-350m} and \textit{bloom-1b3}. However, those two variants were deprecated over time, and \textit{bloom-560m} and \textit{bloom-1b7} were released. Thus all experiments consist of either of these two pairs.

\begin{table*}[h!]
    \centering
    \begin{tabular}{p{5em}p{20em}p{5em}p{5em}}
        \hline
        \textbf{Benchmark/} & \textbf{Dataset} & \multicolumn{2}{c}{\textbf{Evaluated Characteristic}} \\
        \cline{3-4}
        \textbf{Task Type} & & \textbf{Syntactic} & \textbf{Semantic} \\
        \hline
        GLUE & COLA \cite{warstadt-etal-2019-neural} & Correctness & \\ 
        GLUE & QNLI \cite{rajpurkar-etal-2016-squad}, WNLI \cite{levesque2011winograd}, MNLI \cite{bowman-etal-2015-large, williams-etal-2018-broad} & & Entailment\\
        GLUE & MRPC \cite{dolan2005automatically}, STS-B \cite{cer-etal-2017-semeval}, QQP\footnote{\url{https://data.quora.com/First-Quora-Dataset-Release-Question-Pairs}} & & Paraphrase\\
        GLUE, FSL  & SST2 \cite{socher2013recursive} & & Sentiment\\
        GLUE, FSL & RTE \cite{dagan2006pascal,bar2006second,giampiccolo-etal-2007-third,bentivogli2009fifth} & & Entailment\\
        \hline
        QA & SQuAD \cite{rajpurkar-etal-2016-squad} & & QA\\
        \hline
        FSL & AGNews \cite{zhang2015character} & & Topic\\
        FSL & TREC \cite{voorhees2000building} & & QA\\
        FSL & CB \cite{de2019commitmentbank} & & Entailment\\
        FSL & ATIS \cite{hemphill-etal-1990-atis}, MIT Movies \cite{liu2012conversational} & & Entity Ext.\\
        \hline
        FSL, ML & XNLI \cite{conneau-etal-2018-xnli} & & Entailment\\
        ML & MARC \cite{marc_reviews} & &\\
        \hline
        TG, QA & ParaQA \cite{kacupaj2021paraqa}, FiQA \cite{10.1145/3184558.3192301} & & QA\\
        TG & NewsData\footnote{\url{https://www.kaggle.com/datasets/gennadiyr/us-equities-news-data}} & & Topic\\
        \hline
        NER & CoNLL++ \cite{wang-etal-2019-crossweigh} & & Entity Ext.\\
        \hline
        Toxicity & RealToxicityPrompts \cite{gehman-etal-2020-realtoxicityprompts} & & \\
        \hline
    \end{tabular}
    \caption{\label{tab:dataset_list} List of datasets used for evaluation, and their mapping to tasks and evaluated linguistic characteristic as described in Section \ref{sec:motivation}. \textbf{FSL} - Zero-shot and Few-shot Learning, \textbf{QA} - Question Answering, \textbf{ML} - Multi-lingual, \textbf{TG} - Text Generation, \textbf{NER} - Named Entity Recognition.}
\end{table*}

\subsection{Experimental Setup}

For all experiments, models with less than 1B parameters were fine-tuned on 4 NVIDIA Volta (V100) 16-GB GPUs, with a learning rate of $5e-5$ and batch size of 8. For models with more than 1B parameters, 8 NVIDIA Ampere (A100) 32-GB GPUs with AWS Sagemaker\footnote{\url{https://github.com/aws/sagemaker-python-sdk}} Model Parallel, a learning rate of $2e-6$ and batch size of 4 was used for training.

\textbf{GLUE}: All models were fine-tuned on each task's dataset using the \textit{run\_glue.py}\footnote{\url{https://github.com/huggingface/transformers/blob/v4.21.0/examples/pytorch/text-classification/run_glue.py}} script provided in the transformers package \cite{wolf2020transformers}. All evaluations were done on the validation set, as the test set labels have not been released publicly. \textbf{SQuAD}: The experiments were carried out using GPT2sQA\footnote{\url{https://github.com/ftarlaci/GPT2sQA}}. The repository contains code to train a GPT2 model on the SQuAD datasets (2.0 and 1.1). The same codebase was updated to use \textit{bloom-350m/bloom-560m} model. \textbf{Zero- and Few-shot}: We use the codebase released by \citet{zhao2021calibrate} as part of their work on prompt input calibration that was evaluated on GPT-2 and GPT-3. The code was updated to train GPT-Neo and BLOOM models.

\subsection{GLUE}
\label{subsubsec:glue}

Table \ref{tab:glue_results} shows the results of baseline model evaluations on the GLUE tasks. 
DeBERTa has the best performance on six of the nine tasks, and RoBERTa on three. Between the BLOOM models and GPT2, the \textit{gpt2} variant outperforms both BLOOM variants on all tasks. Additionally, for all tasks, \textit{bloom-560m} performs similarly or better than the \textit{bloom-1b7} variant. For the MNLI task, we report the average of the matched and mismatched accuracy scores. The results also show that, except for the MRPC dataset, BLOOM model variants cannot be used in a zero-shot or transfer learning setting. 

\begin{table*}[h!]
    \centering
    \begin{tabular}{p{7em}rrrrrrrrr}
        \hline
        \textbf{Models} & \textbf{COLA} & \textbf{MNLI} & \textbf{MRPC} & \textbf{QNLI} & \textbf{QQP} & \textbf{RTE} & \textbf{SST2} & \textbf{STSB} & \textbf{WNLI}\\
        \hline
        BERT & $0.56$ & $0.84$ & $0.89$ & $0.91$ & $0.88$ & $0.60$ & $0.92$ & $0.89$ & $0.59$\\
        RoBERTa & $0.56$ & $0.87$ & \boldmath$0.93$ & $0.92$ & \boldmath$0.88$ & \boldmath$0.71$ & $0.93$ & $0.90$ & $0.54$\\
        DistilBERT & $0.50$ & $0.81$ & $0.88$ & $0.87$ & $0.86$ & $0.57$ & $0.90$ & $0.86$ & $0.26$\\
        DeBERTa & \boldmath$0.60$ & \boldmath$0.88$ & $0.90$ & \boldmath$0.93$ & $0.88$ & $0.64$ & \boldmath$0.95$ & \boldmath$0.91$ & \boldmath$0.63$\\
        GPT2 & $0.42$ & $0.82$ & $0.86$ & $0.88$ & $0.86$ & $0.68$ & $0.91$ & $0.84$ & $0.57$\\
        BLOOM(560-pt) & $0.01$ & $0.33$ & $0.81$ & $0.50$ & $0.52$ & $0.46$ & $0.51$ & $0.08$ & $0.46$\\
        BLOOM(560-ft) & $0.11$ & $0.70$ & $0.81$ & $0.73$ & $0.82$ & $0.52$ & $0.81$ & $0.41$ & $0.57$\\
        BLOOM(1b7-pt) & $0.04$ & $0.33$ & $0.31$ & $0.48$ & $0.13$ & $0.51$ & $0.49$ & $0.01$ & $0.47$\\
        BLOOM(1b7-ft) & $0.10$ & $0.41$ & $0.81$ & $0.56$ & $0.64$ & $0.52$ & $0.78$ & $0.14$ & $0.56$\\
        \hline
    \end{tabular}
    \caption{\label{tab:glue_results} Fine-tuning result of baselines on GLUE tasks. MNLI scores have been averaged for the matched and mismatched subtasks.}
\end{table*}

For the \textbf{COLA} task, the BLOOM models initially had a Matthews correlation score of 0.0 as all class labels are predicted as acceptable or grammatically correct, resulting in true negatives and false negatives being 0. Similar behavior is observed for the \textbf{SST2} task where the BLOOM models predict the positive class for all samples before and after fine-tuning, resulting in a constant accuracy of 0.509. For the \textbf{MNLI} task, analyzing the predictions showed that the BLOOM models entirely miss/ignore the \textit{neutral} class in \textit{contradiction, entailment, and neutral}. Changing the padding strategy during training helped overcome this behavior and improve model performance.

\subsection{SQuAD}
\label{subsubsec:squad_qa}

The main difference between the two SQuAD versions is that version 2.0 contains questions without answers. The table shows that the \textit{bloom-560m} model outperforms \textit{gpt2-medium} on the SQUAD v.1.1 dataset. Although the metric scores are significantly lower compared to BERT ($\sim$73\% exact match), the results show that when compared to \textit{gpt2-medium}, a similar architecture style model, \textit{bloom-560m} performs better. In the current task setup, the input is formulated as \textit{`[CLS]context[SEP]question[SEP]'} and for questions with no answer, the preprocessing maps the start and end answer index to the `[CLS]' token index. The `[CLS]' token was introduced as part of the input setup for fine-tuning the GPT model \cite{radford2018improving}. However, the `[CLS]' is not part of the BLOOM training or fine-tuning process, making it an unknown token. Although the `[SEP]' and '</s>' tokens are used for different purposes, we hypothesize that replacing the `[CLS]' and `[SEP]' tokens with '<s>' and '</s>,' respectively, will improve the model's performance on no answer questions. We evaluate this by training a \textit{bloom-560m} model (bloom$_R$) with the new input formulation and report the scores in Table \ref{tab:squad_results}.

\begin{table}[]
    \centering
    \begin{tabular}{lllll}
        \hline
        \textbf{Model} & \multicolumn{2}{c}{HA} & \multicolumn{2}{c}{NA}\\
        \cline{2-5}
        & \textbf{EM} & \textbf{F1} & \textbf{EM} & \textbf{F1}\\
        \hline
        \multicolumn{5}{c}{\textbf{SQuAD v2.0}}\\
        \hline
        gpt2 & $5.7$ & $7.19$ & \boldmath$86.96$ & \boldmath$86.96$\\ 
        bloom & \boldmath$10.89$ & \boldmath$15.14$ & $72.58$ & $72.58$\\ 
        bloom$_R$ & $10.81$ & $14.85$ & $72.54$ & $72.54$\\ 
        \hline
        \multicolumn{5}{c}{\textbf{SQuAD v1.1}}\\
        \hline
        gpt2 & $19.75$ & $29.46$ & - & -\\ 
        bloom & \boldmath$24.38$ & \boldmath$33.93$ & - & -\\
        \hline
    \end{tabular}
    \caption{\label{tab:squad_results} Results of gpt2 (\textit{gpt2-medium}) vs bloom (\textit{bloom-560m}) on SQUAD datasets. \textbf{HA} - Questions having an answer, \textbf{NA} - Questions having no answer}
\end{table}

\subsection{Zero- and Few-shot learning}
\label{subsubsec:zero_few_shot}

\begin{figure*}[h!]
    \begin{center}
        \includegraphics[width=0.9\textwidth]{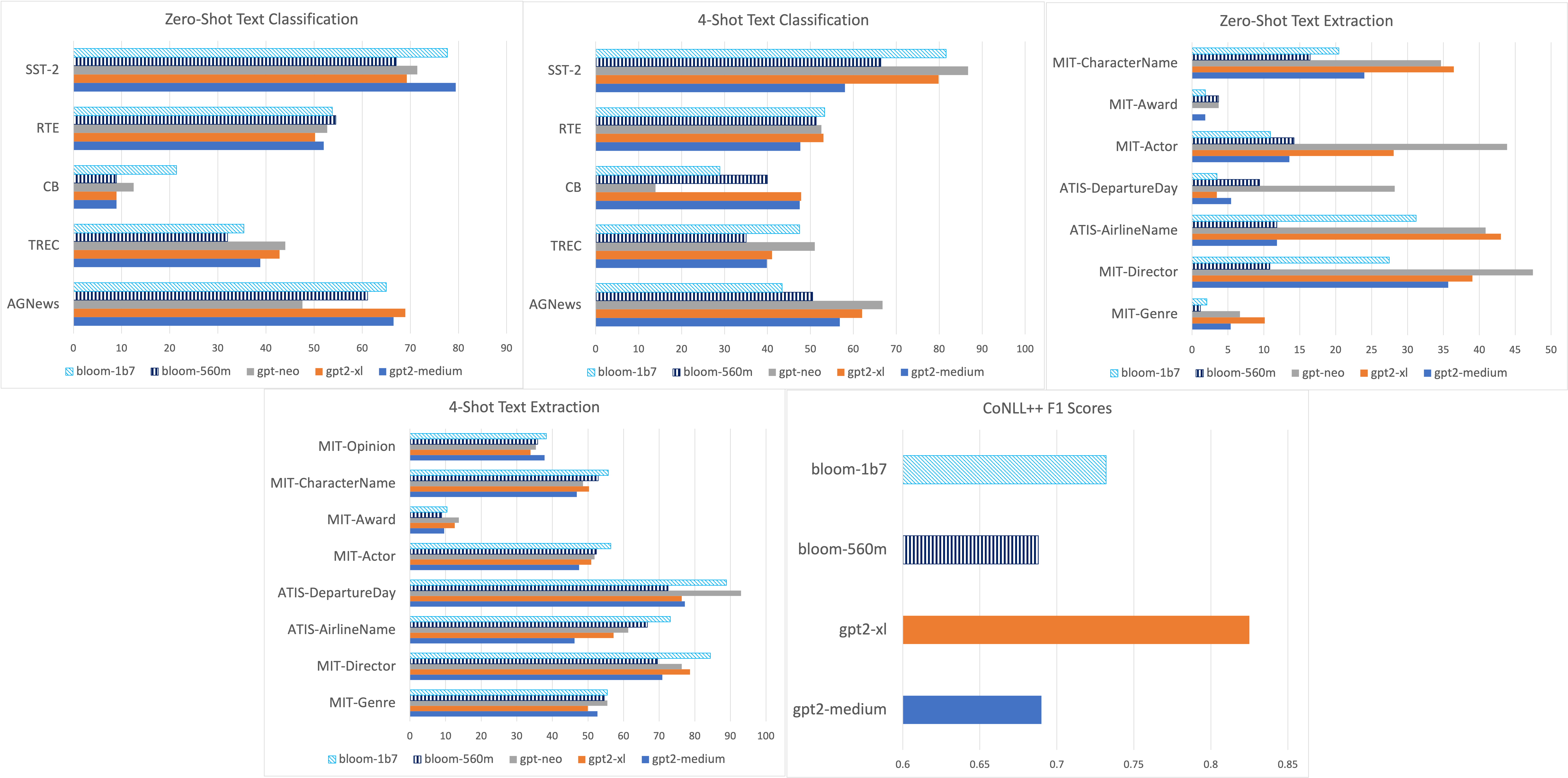}
    \end{center}
    \caption{Zero-shot and four-shot evaluation of BLOOM variants vs. GPT-style models on text classification and extraction tasks. The bottom right graph shows performance on the CoNLL++ dataset.}
    \label{fig:few_shot_results_comparison}
\end{figure*}

The experiments for zero-shot and few-shot (one-, four-, and eight-shot) learning were carried out on five text classification datasets and extraction of 13 entities from the ATIS and MIT Movies datasets. Although all baselines were evaluated in all four settings, for brevity, we only compare zero-shot and four-shot results in Figure \ref{fig:few_shot_results_comparison}. Additionally, the graphs only show the comparison for eight entity extraction tasks. All baselines report a score of 0.0 for the zero-shot MIT-Opinion extraction task. The results show that in the zero-shot setting, the GPT-style models do better than the BLOOM variants on 14 of the 18 total tasks. However, when given more samples, BLOOM model variants outperform GPT-style models on 10 of the 18 tasks in four- and eight-shot settings. Comparing BLOOM model variants with GPT-style models, BLOOM models, on average, do better at text extraction tasks with an average increase in the accuracy by 0.8\%, 1.18\%, and 0.6\% for one-, four- and eight-shot settings. 

We also evaluate the impact of prompt-calibration to understand the sensitivity of the BLOOM models to the samples and the order in the given input prompt. We compute the average increase in accuracy observed post prompt-calibration for each task and observe that except for AGNews and TREC datasets, BLOOM model variants show a marginal increase in accuracy with 0.16\%-3.82\% for \textit{bloom-560m} and 0.17\%-1.56\% for \textit{bloom-1b7}. For three tasks, prompt-calibration hurts the models' performance, with the most significant decrease of 13.66\% observed for \textit{bloom-1b7} on the CB dataset. 

Furthermore, we evaluate the impact of providing a single sample as input in the prompt with the rise in model accuracy in one-shot experiments. The comparison is done by computing the average increase in accuracy across all text classification and text extraction tasks, respectively. The \textit{bloom-560m} variant shows the highest average increase in the accuracy of 22.67\% for text extraction tasks compared to 12.78\% exhibited by the best-performing GPT-style model. The same is not valid for text classification tasks where GPT-Neo shows the highest average increase in the accuracy of 6.38\% post seeing a single sample in the input prompt. Also, it takes a minimum of four samples in the input prompt for the \textit{bloom-1b7} variant to show a positive average increase in accuracy compared to the zero-shot performance.

\citet{scao2022bloom} also evaluate the BLOOM model on the CB and RTE datasets in a prompt-based zero- and few-shot text classification setting. However, the authors do not report the scores for the RTE dataset. Additionally, they only evaluate the 176B BLOOM model for zero-shot settings. For one-shot, they report the performance of all BLOOM variants on the CB dataset. Although the reported scores for 560m and 1b7 models are higher than those reported for 560m and 1b7, it is important to note that they use a set of 5 prompts compared to a single prompt used in this work.

The performance of the BLOOM models on the text extraction tasks led us to evaluate the named entity recognition task for a holistic evaluation of entity extraction. We evaluate the GPT-2 and BLOOM model variants on the CoNLL++ NER dataset. The results of the experiments are displayed as a graph in Figure \ref{fig:few_shot_results_comparison} and show that the GPT-2 model variants outperform the BLOOM model variants. However, similar to results on the text extraction tasks, \textit{bloom-1b7} exhibits better scores over \textit{bloom-560m} for the NER task, thereby indicating a possible area where the larger variant can be used.

\subsection{Multilingual}
\label{subsubsec:multi_lingual}

In multilingual evaluation, we use the XNLI corpus to do cross-language evaluation by training on English and zero-shot evaluation on German. The motivation to pick German as the testing language is twofold - (i) \citet{scao2022bloom} evaluate a fine-tuned variant of BLOOM on various languages in the XNLI dataset in a zero-shot manner. However, the authors do not evaluate the fine-tuned model on German. (ii) The corpus used for pretraining BLOOM does not contain text in the German language, making the evaluation completely zero-shot. Table \ref{tab:xnli_task_scores} shows the baseline model results on the XNLI validation set. The \textit{xlm-roberta-base} model performs the best with an accuracy of 76.15\%, and \textit{bloom-560m} model outperforms the \textit{bloom-1b7} model with and without fine-tuning. 

We divide the evaluated baselines into groups - those with German text in the pretraining corpus (BERT, RoBERTa, and DistilBERT), and those without German text in the pretraining corpus (GPT2 and BLOOM). The results show that knowing the language results in better zero-shot evaluation performance. The second group of the baselines, however, is the important one. GPT2, being pretrained on only English text, still outperforms the BLOOM models that are pretrained on 46 natural languages. Even with fine-tuning on the XNLI English training dataset, the BLOOM models do only slightly better than the random baseline (33\%). This leads to a key conclusion - increasing the number of languages in the training corpus does not necessarily result in improved performance on an unseen language.

\begin{table}[h!]
    \centering
    \begin{tabular}{lr}
        \hline
        \textbf{Models} & \textbf{de}\\
        \hline
        bert-base-multilingual-cased & $71.86$\\
        xlm-roberta-base & \boldmath$76.15$\\
        distilbert-base-multilingual-cased & $65.59$\\
        gpt2-medium & $40.6$\\
        bloom-560m-pt & $33.53$\\
        bloom-560m-ft & $37.75$\\
        bloom-1b7-pt & $33.37$\\
        bloom-1b7-ft & $34.7$\\
        \hline
    \end{tabular}
    \caption{\label{tab:xnli_task_scores} Fine-tuning results for multilingual baselines on XNLI task for zero-shot evaluation on XNLI-de.}
\end{table}

Next, we evaluate the BLOOM model variants using multilingual training and evaluating using the MARC dataset. The MARC dataset, similar to XNLI, contains two languages (German and Japanese) that are not part of the BLOOM model's pretraining corpus. However, unlike the XNLI evaluation, we train the model on instances from all languages here. Table \ref{tab:marc_task_scores} shows the overall accuracy scores along with scores on the German (de) and Japanese (ja) sections of the test set. The results show three important things - (1) the smaller variants perform better than the larger ones, (2) the \textit{bloom-560m} model, although slightly, outperforms both GPT-2 models, and (3) the GPT models pretrained on just English can exhibit similar or better performance than the equivalent sized BLOOM models pretrained on 46 natural languages. Lastly, we compare the models using the accuracy scores on the English instances of the test set. All baselines have been pretrained on English text, and for these instances, \textit{gpt2-medium} reports +1.32\% improvement in the accuracy scores over \textit{bloom-560m}.

\begin{table}[h!]
    \centering
    \begin{tabular}{lrrrrrrr}
        \hline
        \textbf{Models} & \textbf{Acc.} & \textbf{de} & \textbf{ja}\\
        \hline
        gpt2-medium & $61.30$ & \boldmath$65.26$ & $58.04$\\
        gpt2-xl & $55.25$ & $57.96$ & $47.52$\\
        bloom-560m & \boldmath$61.46$ & $63.46$ & \boldmath$58.28$\\
        bloom-1b7 & $43.43$ & $47.96$ & $40.6$\\
        \hline
    \end{tabular}
    \caption{\label{tab:marc_task_scores} Results of GPT-2 and BLOOM model variants on the MARC dataset. \textbf{Acc.} - Accuracy on the entire test set, \textbf{de} - Accuracy on the German test set, \textbf{ja} - Accuracy on the Japanese test set.}
\end{table}

\subsection{Text generation}
\label{subsubsec:text_gen}

Table \ref{tab:bleu_scores_text_generation} shows the results of evaluating the generated texts against the corresponding original answer, context, or article on a few samples using the BLEU metric. GPT2 models do a better job than BLOOM variants across the three generation experiments, and the \textit{bloom-350m} variant does a better job than \textit{bloom-1b3}. However, it is essential to note that the metric scores do not consider the generation of important keywords related to the prompt compared to generating stopwords. The generated text is compared with the original answer/article, which may require additional domain knowledge.

\begin{table}[h!]
    \centering
    \begin{tabular}{lccc}
        \hline
        \textbf{Model} & \textbf{ParaQA} & \textbf{FiQA} & \textbf{NewsData}\\ 
        & (10$^{-2}$) & (10$^{-4}$) & (10$^{-4}$)\\ 
        \hline
        gpt2-medium & $13.6$ & \boldmath$1.82$ & \boldmath$1.72$\\
        gpt2-xl & $14.3$ & $1.69$ & $1.52$\\
        gpt-neo & \boldmath$14.3$ & $1.40$ & $0.65$\\
        bloom-350m & $14.1$ & $1.62$ & $1.27$\\
        bloom-1b3 & $13.9$ & $1.51$ & $1.12$\\
        \hline
    \end{tabular}
    \caption{BLEU scores of different baselines used for the text generation tasks.}\label{tab:bleu_scores_text_generation}
\end{table}

\subsection{Toxicity}
\label{subsubsec:toxicity}

\begin{table}[]
    \centering
    \begin{tabular}{p{6em}p{3em}p{4em}p{3em}}
        \hline
        \textbf{Model} & \textbf{toxicity} & \textbf{severe toxicity} & \textbf{obscene}\\
        \hline
        gpt2 & $0.313$ & $0.022$ & $0.183$\\
        gpt3 \cite{brown2020language} & $0.331$ & $0.021$ & $0.178$\\
        bloom-1b7 & \boldmath$0.26$ & \boldmath$0.016$ & \boldmath$0.122$\\
        \hline
    \end{tabular}
    \caption{\label{tab:toxicity_results} Results of toxicity analysis on RealToxicityPrompts dataset.}
\end{table}

We test the toxicity of the text generated by \textit{bloom-1b7} when prompted to do so using the RealToxicityPrompts \cite{gehman-etal-2020-realtoxicityprompts} dataset. We use the Detoxify\footnote{\url{https://github.com/unitaryai/detoxify}} package to measure the toxicity of the generated text. The \textit{original} model released as part of the package scores the input text on six aspects - toxicity, severe toxicity, obscene, threat, insult, and identity attack. \citet{gehman-etal-2020-realtoxicityprompts} release the text generated by all evaluated baselines, and we use the same to compute toxicity scores and compare with the \textit{bloom-1b7} model. The 175B parameter variant, \textit{DA VINCI}, is used as the GPT-3 model. The toxicity analysis scores reported are computed by taking the average of all scores for all instances in the dataset (Table \ref{tab:toxicity_results}). The results show that when given a prompt that can lead an LLM to generate toxic text, \textit{bloom-1b7} generates less toxic text than gpt2 and gpt3.

\section{Discussion}
\label{sec:discussion}

In this section, we discuss the results presented in the paper to understand BLOOM  and motivate its use in future use cases. We enable this by addressing the following key questions. 
(1) Why do multilingual BLOOM variants perform similarly or worse than monolingual GPT-2 variants? (2) Why is the \textit{bloom-560m} variant better than \textit{bloom-1b7} on multiple tasks? 

\subsection{Multilingual task performance}


BLOOM models underperform in multilingual scenarios even though the training data includes multilingual text. We hypothesize that this is due to the tokenization effects \cite{impact_of_tokenization}. Monolingual models have been shown to perform better than their multilingual counterparts across encoder- and decoder-only architectures \cite{is_mbert_fluent}. \citet{deepmind_tokenizers} find that monolingual tokenizer significantly outperforms multilingual version in 38 out of 48 tasks.  

\subsection{Model size}

The number of parameters in the model directly impacts the training and evaluation environment for any task, precisely disk space, memory size, and time taken to train/evaluate. While numerous large-scale models have been made available as web APIs (GPT-3, BLOOM-176B, ChatGPT), they have limited free access. Thus, we focus on an extensive evaluation of the smaller BLOOM variants that are often favored due to practicality. The experiments carried out as part of this work show that for most of the fine-tuning tasks, the \textit{bloom-560m} variant does a similar or better job when compared to \textit{bloom-1b7}. We explore if undertraining or hyperparameter tuning is attributing to this behavior. 

We use two separate tasks that are different in nature and dataset size - COLA and CoNLL++. For COLA, we conduct multiple experiments with different learning rates, batch sizes, and learning rate schedulers. For all these experiments, we increase the number of epochs to 100 with an early stopping of 5 epochs. However, these experiments do not result in a positive change in the metric scores. For the NER task using CoNLL++, the experiments discussed before had the \textit{bloom-1b7} model trained on five epochs. We train the same model for ten epochs and observe an increase of 0.0027 F1 points. However, the marginal increase in the accuracy fails to establish that undertraining is a probable reason for the poor performance of the \textit{bloom-1b7} model. \citet{xia2022training} analyze this behavior for variants of the OPT model \cite{zhang2022opt} and observe that perplexity is more predictive of model behaviors than model size or training computation. Additionally, the training data distribution can be another reason for this behavior, and we aim to investigate these directions in future work.

Future work may include a better understanding of BLOOM vis-a-vis other LLMs by exploring these questions further. What exactly makes BLOOM models favor text extraction tasks? Why do they do better than GPT-2  on SQuAD questions with answers? Compared to GPT-style models, why do BLOOM models underperform in a zero-shot setting? Why does adding a single instance help BLOOM models more than GPT-style models? and Why does \textit{bloom-1b7} generate less toxic text?  

\section{Conclusion}
\label{sec:conclusion}

The paper empirically evaluates the \textit{bloom-350m/560m} and \textit{bloom-1b3/1b7} variants of the recently released BLOOM model on a diverse set of NLP tasks. The empirical results show that the 560m variant performs at par or better than 1b7 on fine-tuning experiments (except NER). For entity extraction tasks in fine-tuning and prompt-based zero-/few-shot settings, the 1b7 model performs better than the 560m. Furthermore, for prompt-based few-shot text extraction tasks, the 1b7 model performs better than GPT-2 and GPT-Neo variants. However, BLOOM variants fail to replicate the same behavior in a zero-shot setting and underperform compared to the GPT baselines. BLOOM variants perform poorly on an unseen language for multilingual tasks, and even post fine-tuning on the unseen language fail to do better than the monolingual GPT-2 models. On GLUE benchmark tasks, the BLOOM models underperform compared to BERT-style and GPT-style models. Lastly, we observe that compared to GPT-style models, BLOOM variants are more robust to the order of inputs in the prompt and generate less toxic text. The experiment results lead us to pose important questions for future work to facilitate a deeper understanding of the BLOOM model family.

\bibliography{anthology,custom}
\bibliographystyle{acl_natbib}

\appendix

\section{Additional Details}
\label{sec:appendix_addn_details}

\subsection{Baselines}

\begin{enumerate}

    \item \textbf{BERT} - Bidirectional Encoder Representations from Transformers (BERT) \cite{devlin-etal-2019-bert} is a model pre-trained on a large English corpus with two objectives - Masked Language Modeling and Next Sentence Prediction. The training process is carried out in a self-supervised fashion.
    
    \item \textbf{RoBERTa}: \citet{DBLP:journals/corr/abs-1907-11692} presented a study investigating BERT pre-training that measures the impact of key hyperparameters and training data size.
    
    \item \textbf{DistilBERT}: DistilBERT \cite{Sanh2019DistilBERTAD} is a transformers model, smaller and faster than BERT, which was pre-trained on the same corpus in a self-supervised fashion, using the BERT base model as a teacher.
    
    \item \textbf{DeBERTa}: DeBERTa \cite{he2021deberta} improves the BERT and RoBERTa models using disentangled attention and enhanced mask decoder. In DeBERTa V3 \cite{he2021debertav3}, the efficiency of DeBERTa the further improved using ELECTRA-Style \cite{clark2020electra} pre-training with Gradient Disentangled Embedding Sharing. 

    \item \textbf{GPT-style}: GPT-2 \cite{radford2019language} is a Transformers model pre-trained on the English language using the Causal Language Modeling objective. Two GPT-2 variants have been used as baselines - \textit{gpt2-medium} and \textit{gpt2-xl}. For all tasks, unless the variant is explicitly mentioned, \textit{GPT-2} refers to the \textit{gpt2-medium} variant of the model.
    
    \item \textbf{GPT-Neo}: GPT-Neo \cite{gpt-neo} is a Transformer based model designed using EleutherAI's replication of the GPT-3 architecture, and it was trained on Pile \cite{gao2020pile}. This baseline is used only for text-generation experiments, and for the same, the \textit{gpt-neo-1.3B} variant is used.
\end{enumerate}

\subsection{Datasets}

\subsubsection{GLUE}
GLUE \cite{wang2019glue}, the General Language Understanding Evaluation benchmark\footnote{\url{https://gluebenchmark.com/}} consists of various tasks that can aid in evaluating natural language understanding systems. In this study, we consider the following tasks - Corpus of Linguistic Acceptability (CoLA) \cite{warstadt-etal-2019-neural}, Microsoft Research Paraphrase Corpus (MRPC) \cite{dolan2005automatically}, Stanford Question Answering Dataset (QNLI) \cite{rajpurkar-etal-2016-squad}, Quora Question Pairs2 dataset (QQP)\footnote{\url{https://data.quora.com/First-Quora-Dataset-Release-Question-Pairs}}, Recognizing Textual Entailment (RTE) \cite{dagan2006pascal,bar2006second,giampiccolo-etal-2007-third,bentivogli2009fifth}, Stanford Sentiment Treebank (SST2) \cite{socher2013recursive}, Semantic Textual Similarity Benchmark (STS-B) \cite{cer-etal-2017-semeval}, Winograd NLI (WNLI) \cite{levesque2011winograd}, and Multi-Genre Natural Language Inference Corpus (MNLI) \cite{bowman-etal-2015-large, williams-etal-2018-broad}. Table \ref{tab:glue_task} provides more information about each task.

\begin{table}[h!]
    \centering
    \begin{tabular}{lp{6em}p{3em}p{4em}}
        \hline
        \textbf{Task} & \textbf{NLP Task} & \textbf{Dataset size} & \textbf{Metric}\\
        \hline
        COLA & Classifying a sentence as being grammatically correct or not & 9,594 & Matthews Correlation\\
        QNLI & Classifying a sentence as containing an answer to the given question or not & 110,206 & Accuracy\\
        MRPC & Paraphrase Detection & 4,076 & F1 score\\
        STS-B & Paraphrase Detection & 7,249 & Pearson-Spearman Correlation\\
        QQP & Paraphrase Detection & 404,276 & F1 score\\
        SST2 & Sentiment Classification & 68,221 & Accuracy\\
        RTE & Entailment & 2,767 & Accuracy\\
        WNLI & Entailment & 706 & Accuracy\\
        MNLI & Entailment & 412,349 & Accuracy\\
        \hline
    \end{tabular}
    \caption{\label{tab:glue_task} Additional information for each of the GLUE tasks. Dataset size is the sum of train and validation set size)}
\end{table}

\subsubsection{Cross-lingual Natural Language Inference (XNLI) Corpus}\label{subsubsec:xnli}

The Cross-lingual Natural Language Inference (XNLI) \cite{conneau-etal-2018-xnli} corpus contains 5,000 test and 2,500 validation pairs for the MultiNLI corpus \cite{williams-etal-2018-broad}. The collection was crowd-sourced and currently serves as additional test and development data for the MultiNLP corpus. The 14 languages comprise \textit{French, Spanish, German, Greek, Bulgarian, Russian, Turkish, Arabic, Vietnamese, Thai, Chinese, Hindi, Swahili} and \textit{Urdu}.

\subsubsection{The Multilingual Amazon Review Corpus (MARC)}\label{subsubsec:marc}

MARC \cite{marc_reviews} is a dataset containing user reviews on products sold on Amazon, and the label is the number of stars, out of five, that the user gave the product. The reviews are in English, Japanese, German, French, Chinese, and Spanish and were collected between November 1, 2015, to November 1, 2019. The corpus is balanced across all classes and languages, with each language having 200,000, 5,000, and 5,000 instances in the training, development, and test sets. Each record in the dataset contains the review text, the review title, the star rating, an anonymized reviewer ID, an anonymized product ID, and the coarse-grained product category (e.g., ‘books’, ‘appliances’, etc.). For the scope of this task, only review text and title are used since they capture the main content of the review.

\subsubsection{Zero-shot and Few-shot Learning}
\label{sec:appendix_zero_few}

Table \ref{tab:few_shot_dataset_description} describes the datasets used for these experiments. Additionally, Table \ref{tab:few_shot_prompt_description} shows the prompt template and example for few tasks in the zero- and few-shot settings.

\begin{table}[h!]
    \centering
    \begin{tabular}{p{5em}p{8em}p{4em}}
        \hline
        \textbf{Dataset} & \textbf{Description} & \textbf{Dataset size}\\
        \hline
        \multicolumn{3}{c}{Text Classification}\\
        \hline
        AGNews \cite{zhang2015character} & 4-way Topic Classification & 7600\\
        TREC \cite{voorhees2000building} & 6-way Question Classification & 500\\
        CB \cite{de2019commitmentbank} & 3-way Textual Entailment Classification & 56\\
        RTE & 2-way Textual Entailment Classification & 277\\
        SST-2 & Sentiment Analysis & 1821\\
        \hline
        \multicolumn{3}{c}{Text Extraction}\\
        \hline
        MIT-G & Movie Genre in MIT Movies \cite{liu2012conversational} & 780\\
        MIT-D & Movie Director in MIT Movies & 415\\
        ATIS-A & Airline in ATIS \cite{hemphill-etal-1990-atis} & 93\\
        ATIS-D & Departure Date in ATIS & 202\\
        \hline
    \end{tabular}
    \caption{\label{tab:few_shot_dataset_description} Additional information for each of the Few shot learning tasks.}
\end{table}

\begin{table*}[h!]
    \centering
    \begin{tabular}{p{5em}p{12em}p{16em}p{4em}}
        \hline
        \textbf{Dataset} & \textbf{Prompt} & \textbf{Example} & \textbf{Answer}\\
        \hline
        AGNews & Classify the news articles into the categories of World, Sports, Business, and Technology. Article: <article> Answer: & Classify the news articles into the categories of World, Sports, Business, and Technology. Article: Fears for T N pension after talks. Unions representing workers at Turner Newall say they are `disappointed' after talks with stricken parent firm Federal Mogul. Answer: & Business\\
        TREC & Classify the questions based on whether their answer type is a Number, Location, Person, Description, Entity, or Abbreviation. Question: <question> Answer Type: & Classify the questions based on whether their answer type is a Number, Location, Person, Description, Entity, or Abbreviation. Question: How far is it from Denver to Aspen? Answer Type: & Number\\
        CB & <premise> question: <hypothesis>. true, false, or neither? answer: & Valence the void-brain, Valence the virtuous valet. Why couldn't the figger choose his own portion of titanic anatomy to shaft? Did he think he was helping? question: Valence was helping. true, false, or neither? answer: & false\\
        RTE & <premise> question: <hypothesis>. True or False? answer: & Dana Reeve, the widow of the actor Christopher Reeve, has died of lung cancer at age 44, according to the Christopher Reeve Foundation. question: Christopher Reeve had an accident. True or False? answer: & False\\
        SST-2 & Review: <input sentence> Sentiment: & Review: gangs of new york is an unapologetic mess , whose only saving grace is that it ends by blowing just about everything up . Sentiment: & Negative\\
        MIT-G & Sentence: <input sentence> Genre: & Sentence: what soviet science fiction classic about a mysterious planet was later remade by steven soderbergh and george clooney Genre: & soviet science fiction\\
        MIT-D & Sentence: <input sentence> Director: & Sentence: what soviet science fiction classic about a mysterious planet was later remade by steven soderbergh and george clooney Director: & steven soderbergh\\
        ATIS-A & Sentence: <input sentence> Airline name: & Sentence: i would like a flight from orlando to salt lake city for april first on delta airlines Airline name: & delta airlines\\
        ATIS-D & Sentence: <input sentence> Depart date - Day name: & Sentence: monday morning i would like to fly from columbus to indianapolis Depart date - Day name: & monday\\
        \hline
    \end{tabular}
    \caption{\label{tab:few_shot_prompt_description} Prompt template and example for few of the zero- and few-shot learning tasks.}
\end{table*}

\section{Additional Results}
\label{sec:appendix_addn_results}

\begin{table*}
    \centering
    \begin{tabular}{llrrrr}
        \hline
        \textbf{Dataset} & \textbf{Models} & \textbf{0-shot} & \textbf{1-shot} & \textbf{4-shot} & \textbf{8-shot}\\
        \hline
        AGNews & gpt2-medium & 66.48$_{0.0}$ & 65.26$_{4.73}$ & 56.85$_{3.11}$ & 49.14$_{6.68}$\\
        & gpt2-xl & \textbf{68.94}$_{0.0}$ & 66.49$_{3.90}$ & 62.02$_{9.96}$ & 60.24$_{12.63}$\\
        & gpt-neo & 47.57$_{0.0}$ & 52.45$_{5.75}$ & \textbf{66.79}$_{9.31}$ & 67.66$_{8.52}$\\
        & bloom-560m & 61.13$_{0.0}$ & 53.29$_{4.17}$ & 50.51$_{10.50}$ & 43.37$_{6.04}$\\
        & bloom-1b7 & 64.97$_{0.0}$ & \textbf{70.16}$_{5.58}$ & 62.51$_{7.01}$ & \textbf{68.59}$_{9.93}$\\
        \cline{2-6}
        TREC & gpt2-medium & 38.8$_{0.0}$ & 31$_{3.76}$ & 39.84$_{3.82}$ & 39.64$_{2.95}$\\
        & gpt2-xl & 42.80$_{0.0}$ & 40.52$_{3.69}$ & 41.08$_{4.98}$ & 45.62$_{1.86}$\\
        & gpt-neo & \textbf{44.00}$_{0.0}$ & \textbf{57.68}$_{5.99}$ & \textbf{51.04}$_{3.69}$ & 52.16$_{5.19}$\\
        & bloom-560m & 32$_{0.0}$ & 34.88$_{5.38}$ & 35$_{5.54}$ & 42.68$_{8.69}$\\
        & bloom-1b7 & 35.40$_{0.0}$ & 46.48$_{3.31}$ & 47.48$_{11.30}$ & \textbf{57.28}$_{5.73}$\\
        \cline{2-6}
        CB & gpt2-medium & 8.92$_{0.0}$ & \textbf{40.71}$_{7.92}$ & 47.49$_{4.65}$ & 50.44$_{1.70}$\\
        & gpt2-xl & 8.92$_{0.0}$ & 25$_{13.55}$ & \textbf{47.86}$_{8.33}$ & \textbf{53.56}$_{11.57}$\\
        & gpt-neo & 12.50$_{0.0}$ & 15.36$_{10.26}$ & 13.93$_{3.46}$ & 18.57$_{8.86}$\\
        & bloom-560m & 8.93$_{0.0}$ & 29.64$_{16.93}$ & 40$_{15.43}$ & 42.14$_{16.39}$\\
        & bloom-1b7 & \textbf{21.42}$_{0.0}$ & 16.42$_{10.19}$ & 28.92$_{14.26}$ & 29.29$_{13.49}$\\
        \cline{2-6}
        RTE & gpt2-medium & 51.98$_{0.0}$ & 49.02$_{2.15}$ & 47.65$_{0.62}$ & 47.65$_{0.62}$\\
        & gpt2-xl & 50.18$_{0.0}$ & 52.49$_{1.93}$ & 53.06$_{1.16}$ & \textbf{53.42}$_{1.16}$\\
        & gpt-neo & 52.71$_{0.0}$ & 49.68$_{2.20}$ & 52.56$_{2.96}$ & 52.13$_{3.71}$\\
        & bloom-560m & \textbf{54.51}$_{0.0}$ & \textbf{52.85}$_{3.10}$ & 51.34$_{3.55}$ & 53$_{4.32}$\\
        & bloom-1b7 & 53.79$_{0.0}$ & 50.76$_{2.45}$ & \textbf{53.29}$_{3.73}$ & 51.62$_{4.41}$\\
        \cline{2-6}
        SST-2 & gpt2-medium & \textbf{79.40}$_{0.0}$ & 60.91$_{11.58}$ & 58.02$_{14.73}$ & 63.99$_{13.12}$\\
        & gpt2-xl & 69.24$_{0.0}$ & 76.40$_{2.48}$ & 79.84$_{11.46}$ & 73.89$_{12.90}$\\
        & gpt-neo & 71.44$_{0.0}$ & \textbf{84.95}$_{3.61}$ & \textbf{86.71}$_{3.81}$ & \textbf{85.35}$_{2.69}$\\
        & bloom-560m & 67.05$_{0.0}$ & 63.01$_{4.95}$ & 66.39$_{5.54}$ & 69.54$_{8.77}$\\
        & bloom-1b7 & 77.70$_{0.0}$ & 64.74$_{5.96}$ & 81.57$_{8.19}$ & 83.61$_{6.09}$\\
        \hline
    \end{tabular}
    \caption{\label{tab:few-shot-text-classification} Results of few-shot learning experiments all the baselines on text-classification tasks.}
\end{table*}

\begin{table*}
    \centering
    \begin{tabular}{llrrrr}
        \hline
        \textbf{Dataset} & \textbf{Models} & \textbf{0-shot} & \textbf{1-shot} & \textbf{4-shot} & \textbf{8-shot}\\
        \hline
        MIT-Genre & gpt2-medium & 5.38$_{0.0}$ & 35.22$_{5.88}$ & 52.66$_{2.89}$ & 55.95$_{2.73}$\\
        & gpt2-xl & \textbf{10.12}$_{0.0}$ & 31.15$_{8.41}$ & 49.91$_{6.50}$ & 54.22$_{6.58}$\\
        & gpt-neo & 6.67$_{0.0}$ & \textbf{44.10}$_{4.07}$ & \textbf{55.46}$_{3.14}$ & \textbf{56.92}$_{3.62}$\\
        & bloom-560m & 1.15$_{0.0}$ & 37.30$_{6.13}$ & 54.58$_{2.26}$ & 56.12$_{1.64}$\\
        & bloom-1b7 & 2.05$_{0.0}$ & 41.22$_{7.41}$ & 55.43$_{3.00}$ & 56.27$_{7.02}$\\
        & bloom-3b & 0.12$_{0.0}$ & - & - & -\\
        \cline{2-6}
        MIT-Director & gpt2-medium & 35.66$_{0.0}$ & 55.65$_{3.70}$ & 70.88$_{2.45}$ & 77.05$_{1.19}$\\
        & gpt2-xl & 39.03$_{0.0}$ & 61.49$_{3.58}$ & 78.69$_{1.29}$ & 81.58$_{0.92}$\\
        & gpt-neo & \textbf{47.47}$_{0.0}$ & 71.13$_{3.98}$ & 76.34$_{2.67}$ & 77.54$_{1.52}$\\
        & bloom-560m & 10.84$_{0.0}$ & 58.64$_{7.60}$ & 69.53$_{5.63}$ & 75.36$_{3.08}$\\
        & bloom-1b7 & 27.46$_{0.0}$ & \textbf{77.87}$_{4.50}$ & \textbf{84.38}$_{2.16}$ & \textbf{83.75}$_{2.24}$\\
        & bloom-3b & 16.86$_{0.0}$ & - & - & -\\
        \cline{2-6}
        MIT-Actor & gpt2-medium & 13.56$_{0.0}$ & 47.99$_{3.69}$ & 47.48$_{2.18}$ & 48.09$_{0.80}$\\
        & gpt2-xl & 28.07$_{0.0}$ & 49.07$_{1.54}$ & 50.93$_{1.79}$ & 51.93$_{1.13}$\\
        & gpt-neo & \textbf{43.86}$_{0.0}$ & 51.29$_{2.42}$ & 51.84$_{1.27}$ & 52.12$_{1.57}$\\
        & bloom-560m & 14.19$_{0.0}$ & 51.46$_{1.65}$ & 52.37$_{0.48}$ & 52.44$_{0.50}$\\
        & bloom-1b7 & 10.91$_{0.0}$ & \textbf{54.56}$_{1.84}$ & \textbf{56.40}$_{0.42}$ & \textbf{56.12}$_{0.64}$\\
        & bloom-3b & 11.22$_{0.0}$ & - & - & -\\
        \cline{2-6}
        MIT-Award & gpt2-medium & 1.85$_{0.0}$ & 5.93$_{4.44}$ & 9.63$_{5.02}$ & 13.70$_{4.77}$\\
        & gpt2-xl & 0.0$_{0.0}$ & 3.70$_{3.88}$ & 12.59$_{4.29}$ & 12.96$_{5.62}$\\
        & gpt-neo & \textbf{3.70}$_{0.0}$ & \textbf{10.37}$_{7.55}$ & \textbf{13.70}$_{3.99}$ & \textbf{15.19}$_{6.35}$\\
        & bloom-560m & \textbf{3.70}$_{0.0}$ & 5.56$_{3.51}$ & 8.89$_{3.51}$ & 11.85$_{4.48}$\\
        & bloom-1b7 & 1.85$_{0.0}$ & 6.30$_{5.57}$ & 10.37$_{6.48}$ & 14.44$_{6.97}$\\
        & bloom-3b & 0.0$_{0.0}$ & - & - & -\\
        \cline{2-6}
        MIT-CharacterName & gpt2-medium & 24.00$_{0.0}$ & 42.49$_{2.36}$ & 46.84$_{4.60}$ & 47.47$_{4.39}$\\
        & gpt2-xl & \textbf{36.44}$_{0.0}$ & 46.04$_{9.61}$ & 50.31$_{3.75}$ & 51.73$_{4.45}$\\
        & gpt-neo & 34.67$_{0.0}$ & 45.07$_{6.76}$ & 48.62$_{5.54}$ & 50.31$_{3.14}$\\
        & bloom-560m & 16.44$_{0.0}$ & \textbf{50.93}$_{1.92}$ & 52.89$_{4.37}$ & 54.21$_{2.68}$\\
        & bloom-1b7 & 20.44$_{0.0}$ & 44.98$_{5.51}$ & \textbf{55.73}$_{2.15}$ & \textbf{57.78}$_{2.91}$\\
        & bloom-3b & 10.22$_{0.0}$ & - & - & -\\
        \cline{2-6}
        ATIS-A & gpt2-medium & 11.82$_{0.0}$ & 41.50$_{19.45}$ & 46.22$_{18.24}$ & 61.07$_{11.86}$\\
        & gpt2-xl & \textbf{43.01}$_{0.0}$ & 48.16$_{17.62}$ & 57.20$_{14.43}$ & 74.61$_{8.48}$\\
        & gpt-neo & 40.86$_{0.0}$ & 57.20$_{14.22}$ & 61.29$_{16.92}$ & \textbf{78.28}$_{7.68}$\\
        & bloom-560m & 11.82$_{0.0}$ & 49.24$_{7.57}$ & 66.66$_{13.32}$ & 64.94$_{13.44}$\\
        & bloom-1b7 & 31.18$_{0.0}$ & \textbf{68.16}$_{11.39}$ & \textbf{73.11}$_{10.64}$ & 74.25$_{7.95}$\\
        \cline{2-6}
        ATIS-D & gpt2-medium & 5.44$_{0.0}$ & 81.48$_{8.76}$ & 77.22$_{9.91}$ & 79.20$_{9.61}$\\
        & gpt2-xl & 3.46$_{0.0}$ & 75.73$_{4.48}$ & 76.33$_{13.40}$ & 86.92$_{3.14}$\\
        & gpt-neo & \textbf{28.22}$_{0.0}$ & \textbf{93.56}$_{2.32}$ & \textbf{92.97}$_{3.09}$ & 95.25$_{1.99}$\\
        & bloom-560m & 9.40$_{0.0}$ & 70.59$_{9.43}$ & 72.47$_{18.47}$ & 73.16$_{12.63}$\\
        & bloom-1b7 & 3.46$_{0.0}$ & 85.44$_{11.38}$ & 88.90$_{15.63}$ & \textbf{95.73}$_{4.94}$\\
        \hline
    \end{tabular}
    \caption{\label{tab:few-shot-text-extraction-part-1} Results of few-shot learning experiments all the baselines on text-extraction tasks.}
\end{table*}

\begin{table*}
    \centering
    \begin{tabular}{llrrrr}
        \hline
        \textbf{Dataset} & \textbf{Models} & \textbf{0-shot} & \textbf{1-shot} & \textbf{4-shot} & \textbf{8-shot}\\
        \hline
        MIT-Opinion & gpt2-medium & 0.0$_{0.0}$ & 7.47$_{6.67}$ & 37.79$_{7.53}$ & 47.47$_{3.67}$\\
        & gpt2-xl & 0.0$_{0.0}$ & 16.84$_{14.77}$ & 33.89$_{13.96}$ & 48.21$_{3.64}$\\
        & gpt-neo & 0.0$_{0.0}$ & \textbf{22.21}$_{13.26}$ & 35.37$_{12.31}$ & \textbf{52.42}$_{3.63}$\\
        & bloom-560m & 0.0$_{0.0}$ & 15.57$_{15.14}$ & 35.89$_{7.30}$ & 45.25$_{4.80}$\\
        & bloom-1b7 & 0.0$_{0.0}$ & 18.21$_{0.87}$ & \textbf{38.31}$_{12.69}$ & 51.99$_{5.28}$\\
        & bloom-3b & 0.0$_{0.0}$ & - & - & -\\
        \cline{2-6}
        MIT-Origin & gpt2-medium & \textbf{0.56}$_{0.0}$ & 4.04$_{1.15}$ & 4.38$_{1.60}$ & 5.51$_{1.93}$\\
        & gpt2-xl & 0.0$_{0.0}$ & 2.81$_{1.81}$ & 4.83$_{1.80}$ & 6.18$_{2.78}$\\
        & gpt-neo & 0.0$_{0.0}$ & \textbf{4.49}$_{3.14}$ & 5.28$_{3.29}$ & 7.30$_{3.12}$\\
        & bloom-560m & \textbf{0.56}$_{0.0}$ & 3.70$_{2.0}$ & 4.82$_{2.33}$ & \textbf{7.63}$_{2.43}$\\
        & bloom-1b7 & 0.0$_{0.0}$ & 4.26$_{3.47}$ & \textbf{6.17}$_{2.72}$ & 7.41$_{3.06}$\\
        & bloom-3b & 0.0$_{0.0}$ & - & - & -\\
        \cline{2-6}
        MIT-Plot & gpt2-medium & 0.07$_{0.0}$ & 0.05$_{0.05}$ & 1.04$_{0.5}$ & 1.40$_{0.69}$\\
        & gpt2-xl & 0.07$_{0.0}$ & 0.32$_{0.28}$ & \textbf{1.65}$_{0.38}$ & \textbf{1.96}$_{0.71}$\\
        & gpt-neo & 0.07$_{0.0}$ & 0.33$_{0.35}$ & 1.37$_{0.45}$ & \textbf{1.96}$_{0.68}$\\
        & bloom-560m & 0.06$_{0.0}$ & 0.42$_{0.29}$ & 1.62$_{0.48}$ & 1.77$_{0.57}$\\
        & bloom-1b7 & \textbf{0.13}$_{0.0}$ & \textbf{0.51}$_{0.38}$ & 1.57$_{0.96}$ & 1.55$_{0.75}$\\
        & bloom-3b & 0.0$_{0.0}$ & - & - & -\\
        \cline{2-6}
        MIT-Quote & gpt2-medium & 0.0$_{0.0}$ & 6.05$_{3.15}$ & 15.35$_{3.48}$ & 17.21$_{2.37}$\\
        & gpt2-xl & 0.0$_{0.0}$ & 13.94$_{6.57}$ & \textbf{20.46}$_{2.54}$ & 21.39$_{2.54}$\\
        & gpt-neo & \textbf{2.33}$_{0.0}$ & \textbf{14.88}$_{3.15}$ & 20$_{2.37}$ & 19.53$_{3.12}$\\
        & bloom-560m & 0.0$_{0.0}$ & 11.16$_{2.71}$ & 17.67$_{4.31}$ & 20.47$_{1.74}$\\
        & bloom-1b7 & \textbf{2.33}$_{0.0}$ & 13.49$_{5.38}$ & 20$_{3.48}$ & \textbf{21.86}$_{1.86}$\\
        & bloom-3b & 2.32$_{0.0}$ & - & - & -\\
        \cline{2-6}
        MIT-Relationship & gpt2-medium & 0.0$_{0.0}$ & 4.49$_{7.98}$ & 21.22$_{7.27}$ & 31.02$_{6.56}$\\
        & gpt2-xl & 0.0$_{0.0}$ & 8.02$_{5.29}$ & \textbf{32.10}$_{12.43}$ & 31.02$_{10.12}$\\
        & gpt-neo & 0.0$_{0.0}$ & 11.56$_{8.96}$ & 30.61$_{10.81}$ & 29.52$_{12.93}$\\
        & bloom-560m & 0.0$_{0.0}$ & 10.88$_{9.62}$ & 26.26$_{4.80}$ & \textbf{31.16}$_{7.15}$\\
        & bloom-1b7 & 0.0$_{0.0}$ & \textbf{15.92}$_{9.48}$ & 23.67$_{11.72}$ & 27.89$_{13.91}$\\
        & bloom-3b & 0.0$_{0.0}$ & - & - & -\\
        \cline{2-6}
        MIT-Year & gpt2-medium & \textbf{92.37}$_{0.0}$ & 87.79$_{6.24}$ & 93.10$_{1.37}$ & 93.16$_{2.19}$\\
        & gpt2-xl & 89.16$_{0.0}$ & 84.91$_{9.59}$ & 92.14$_{1.51}$ & 94.25$_{1.55}$\\
        & gpt-neo & \textbf{92.37}$_{0.0}$ & 91.27$_{2.48}$ & 93.74$_{1.52}$ & 93.59$_{1.45}$\\
        & bloom-560m & 49.01$_{0.0}$ & 85.04$_{8.58}$ & 91.42$_{2.77}$ & 92.40$_{2.05}$\\
        & bloom-1b7 & 88.09$_{0.0}$ & \textbf{92.40}$_{2.10}$ & \textbf{94.02}$_{1.46}$ & \textbf{94.78}$_{1.10}$\\
        & bloom-3b & 81.67$_{0.0}$ & - & - & -\\
        \hline
    \end{tabular}
    \caption{\label{tab:few-shot-text-extraction-part-2} Results of few-shot learning experiments all the baselines on text-extraction tasks.}
\end{table*}

\begin{table*}
    \centering
    \begin{tabular}{llrrrrr}
        \hline
        \textbf{Dataset} & \textbf{Models} & \textbf{0-shot} & \textbf{1-shot} & \textbf{4-shot} & \textbf{8-shot} & \textbf{Avg. Increase}\\
        \hline
        AGNews & o-bloom-560m & \textbf{63.87}$_{0.0}$ & 30.48$_{6.90}$ & 29.07$_{5.07}$ & 26.19$_{1.55}$ & \\
        & bloom-560m & 61.13$_{0.0}$ & \textbf{53.29}$_{4.17}$ & \textbf{50.51}$_{10.50}$ & \textbf{43.37}$_{6.04}$ & 14.67\\
        & o-bloom-1b7 & 50.03$_{0.0}$ & 36.62$_{12.25}$ & 33.25$_{5.34}$ & 40.03$_{14.28}$ & \\
        & bloom-1b7 & \textbf{64.97}$_{0.0}$ & \textbf{70.16}$_{5.58}$ & \textbf{62.51}$_{7.01}$ & \textbf{68.59}$_{9.93}$ & 26.57\\
        \cline{2-7}
        TREC & o-bloom-560m & 22.60$_{0.0}$ & 23.36$_{7.64}$ & 20.88$_{9.14}$ & 28.84$_{6.76}$ & \\
        & bloom-560m & \textbf{32}$_{0.0}$ & \textbf{34.88}$_{5.38}$ & \textbf{35}$_{5.54}$ & \textbf{42.68}$_{8.69}$ & 12.22\\
        & o-bloom-1b7 & 22.80$_{0.0}$ & 39.16$_{2.33}$ & 29.56$_{14.37}$ & 43.36$_{11.23}$ & \\
        & bloom-1b7 & \textbf{35.40}$_{0.0}$ & \textbf{46.48}$_{3.31}$ & \textbf{47.48}$_{11.30}$ & \textbf{57.28}$_{5.73}$ & 12.94\\
        \cline{2-7}
        CB & o-bloom-560m & \textbf{42.86}$_{0.0}$ & 28.21$_{15.75}$ & 38.57$_{15.47}$ & \textbf{45.71}$_{5.13}$ & \\
        & bloom-560m & 8.93$_{0.0}$ & \textbf{29.64}$_{16.93}$ & \textbf{40}$_{15.43}$ & 42.14$_{16.39}$ & -8.66\\
        & o-bloom-1b7 & \textbf{42.86}$_{0.0}$ & \textbf{28.21}$_{15.75}$ & \textbf{36.79}$_{13.49}$ & \textbf{42.14}$_{2.14}$ & \\
        & bloom-1b7 & 21.42$_{0.0}$ & 16.42$_{10.19}$ & 28.92$_{14.26}$ & 29.29$_{13.49}$ & -13.48\\
        \cline{2-7}
        RTE & o-bloom-560m & 52.71$_{0.0}$ & \textbf{54.15}$_{1.51}$ & \textbf{52.20}$_{2.79}$ & \textbf{53.21}$_{2.27}$ & \\
        & bloom-560m & \textbf{54.51}$_{0.0}$ & 52.85$_{3.10}$ & 51.34$_{3.55}$ & 53$_{4.32}$ & -0.21\\
        & o-bloom-1b7 & 53.43$_{0.0}$ & \textbf{50.83}$_{3.34}$ & 53$_{5.59}$ & \textbf{51.77}$_{3.95}$ & \\
        & bloom-1b7 & \textbf{53.79}$_{0.0}$ & 50.76$_{2.45}$ & \textbf{53.29}$_{3.73}$ & 51.62$_{4.41}$ & 0.10\\
        \cline{2-7}
        SST-2 & o-bloom-560m & \textbf{73.31}$_{0.0}$ & \textbf{67.14}$_{7.06}$ & 59.74$_{9.44}$ & 57.17$_{9.63}$ & \\
        & bloom-560m & 67.05$_{0.0}$ & 63.01$_{4.95}$ & \textbf{66.39}$_{5.54}$ & \textbf{69.54}$_{8.77}$ & 2.15\\
        & o-bloom-1b7 & \textbf{77.98}$_{0.0}$ & \textbf{74.84}$_{8.79}$ & \textbf{87.49}$_{3.49}$ & 73.07$_{11.41}$ & \\
        & bloom-1b7 & 77.70$_{0.0}$ & 64.74$_{5.96}$ & 81.57$_{8.19}$ & \textbf{83.61}$_{6.09}$ & -1.44\\
        \hline
    \end{tabular}
    \caption{\label{tab:few-shot-text-classification-org-vs-cal} Results of few-shot learning experiments on bloom models with and without prompt calibration for text extraction tasks.}
\end{table*}

\begin{table*}
    \centering
    \begin{tabular}{llrrrrr}
        \hline
        \textbf{Dataset} & \textbf{Models} & \textbf{0-shot} & \textbf{1-shot} & \textbf{4-shot} & \textbf{8-shot} & \textbf{Avg. Increase}\\
        \hline
        MIT-Actor & o-bloom-560m & 14.09$_{0.0}$ & 47.19$_{7.05}$ & 50.91$_{1.87}$ & 52.27$_{0.91}$ & \\
        & bloom-560m & \textbf{14.19}$_{0.0}$ & \textbf{51.46}$_{1.65}$ & \textbf{52.37}$_{0.48}$ & \textbf{52.44}$_{0.50}$ & 1.5\\
        & o-bloom-1b7 & 10.70$_{0.0}$ & 52.50$_{3.54}$ & 56.36$_{0.53}$ & 56.12$_{0.64}$ & \\
        & bloom-1b7 & \textbf{10.91}$_{0.0}$ & \textbf{54.56}$_{1.84}$ & \textbf{56.40}$_{0.42}$ & 56.12$_{0.64}$ & 0.57\\
        \cline{2-7}
        MIT-Award & o-bloom-560m & 3.70$_{0.0}$ & 4.07$_{3.19}$ & \textbf{9.26}$_{5.62}$ & 11.11$_{4.06}$ & \\
        & bloom-560m & 3.70$_{0.0}$ & \textbf{5.56}$_{3.51}$ & 8.89$_{3.51}$ & \textbf{11.85}$_{4.48}$ & 0.46\\
        & o-bloom-1b7 & 1.85$_{0.0}$ & 5.19$_{5.79}$ & \textbf{11.85}$_{6.37}$ & 12.22$_{8.00}$ & \\
        & bloom-1b7 & 1.85$_{0.0}$ & \textbf{6.30}$_{5.57}$ & 10.37$_{6.48}$ & \textbf{14.44}$_{6.97}$ & 0.46\\
        \cline{2-7}
        MIT-CharacterName & o-bloom-560m & 16.44$_{0.0}$ & 48.09$_{3.39}$ & 51.56$_{4.88}$ & 52.36$_{4.57}$ & \\
        & bloom-560m & 16.44$_{0.0}$ & \textbf{50.93}$_{1.92}$ & \textbf{52.89}$_{4.37}$ & \textbf{54.22}$_{2.40}$ & 1.50\\
        & o-bloom-1b7 & 20.00$_{0.0}$ & 44.44$_{5.77}$ & 55.73$_{2.17}$ & \textbf{57.96}$_{2.76}$ & \\
        & bloom-1b7 & \textbf{20.44}$_{0.0}$ & \textbf{44.98}$_{5.51}$ & 55.73$_{2.15}$ & 57.78$_{2.91}$ & 0.2\\
        \cline{2-7}
        MIT-Opinion & o-bloom-560m & 0.0$_{0.0}$ & 13.16$_{12.75}$ & 35.79$_{7.16}$ & \textbf{46.11}$_{4.70}$ & \\
        & bloom-560m & 0.0$_{0.0}$ & \textbf{15.58}$_{13.55}$ & \textbf{35.89}$_{6.53}$ & 45.26$_{4.30}$ & 0.41\\
        & o-bloom-1b7 & 0.0$_{0.0}$ & 15.37$_{13.78}$ & \textbf{39.16}$_{11.57}$ & \textbf{52.42}$_{4.77}$ & \\
        & bloom-1b7 & 0.0$_{0.0}$ & \textbf{18.21}$_{0.87}$ & 38.32$_{11.35}$ & 52$_{4.72}$ & 0.39\\
        \cline{2-7}
        MIT-Origin & o-bloom-560m & 0.0$_{0.0}$ & 0.45$_{0.42}$ & 4.16$_{2.18}$ & 6.52$_{1.61}$ & \\
        & bloom-560m & \textbf{0.56}$_{0.0}$ & \textbf{1.69}$_{1.28}$ & \textbf{4.83}$_{2.09}$ & \textbf{7.64}$_{2.18}$ & 0.89\\
        & o-bloom-1b7 & 0.0$_{0.0}$ & 1.35$_{0.98}$ & 4.83$_{2.63}$ & 6.52$_{2.60}$ & \\
        & bloom-1b7 & 0.0$_{0.0}$ & \textbf{4.27}$_{3.11}$ & \textbf{6.18}$_{2.44}$ & \textbf{7.42}$_{2.74}$ & 1.29\\
        \cline{2-7}
        MIT-Plot & o-bloom-560m & 0.07$_{0.0}$ & 0.26$_{0.17}$ & 1.33$_{0.30}$ & 1.77$_{0.64}$ & \\
        & bloom-560m & 0.07$_{0.0}$ & \textbf{0.42}$_{0.27}$ & \textbf{1.63}$_{0.44}$ & \textbf{1.95}$_{0.55}$ & 0.16\\
        & o-bloom-1b7 & 0.07$_{0.0}$ & 0.23$_{0.18}$ & 1.22$_{0.17}$ & 1.51$_{0.67}$ &\\
        & bloom-1b7 & \textbf{0.14}$_{0.0}$ & \textbf{0.52}$_{0.34}$ & \textbf{1.49}$_{0.61}$ & \textbf{1.56}$_{0.68}$ & 0.17\\
        \cline{2-7}
        MIT-Quote & o-bloom-560m & 0.0$_{0.0}$ & 7.44$_{4.74}$ & 17.67$_{4.31}$ & 20.47$_{1.74}$ & \\
        & bloom-560m & 0.0$_{0.0}$ & \textbf{11.16}$_{2.71}$ & 17.67$_{4.31}$ & 20.47$_{1.74}$ & 0.93\\
        & o-bloom-1b7 & 2.33$_{0.0}$ & 11.63$_{4.41}$ & 20$_{3.48}$ & 21.86$_{1.86}$ & \\
        & bloom-1b7 & 2.33$_{0.0}$ & \textbf{13.49}$_{5.38}$ & 20$_{3.48}$ & 21.86$_{1.86}$ & 0.46\\
        \cline{2-7}
        MIT-Relationship & o-bloom-560m & 0.0$_{0.0}$ & 5.03$_{9.05}$ & 23.40$_{6.95}$ & 30.20$_{8.01}$ & \\
        & bloom-560m & 0.0$_{0.0}$ & \textbf{10.88}$_{9.62}$ & \textbf{26.26}$_{4.80}$ & \textbf{31.16}$_{7.15}$ & 2.41\\
        & o-bloom-1b7 & 0.0$_{0.0}$ & 9.25$_{7.67}$ & \textbf{23.81}$_{11.67}$ & \textbf{28.16}$_{13.35}$ & \\
        & bloom-1b7 & 0.0$_{0.0}$ & \textbf{15.92}$_{9.48}$ & 23.67$_{11.72}$ & 27.89$_{13.91}$ & 1.56\\
        \cline{2-7}
        MIT-Year & o-bloom-560m & 48.85$_{0.0}$ & 71.63$_{12.56}$ & 89.71$_{4.59}$ & 91.54$_{2.90}$ & \\
        & bloom-560m & \textbf{49.01}$_{0.0}$ & \textbf{85.04}$_{8.58}$ & \textbf{91.42}$_{2.77}$ & \textbf{92.40}$_{2.05}$ & 3.82\\
        & o-bloom-1b7 & 88.09$_{0.0}$ & 91.15$_{3.74}$ & 94.02$_{1.45}$ & 94.78$_{1.12}$ & \\
        & bloom-1b7 & 88.09$_{0.0}$ & \textbf{92.40}$_{2.10}$ & 94.02$_{1.46}$ & 94.78$_{1.10}$ & 0.31\\
        \hline
    \end{tabular}
    \caption{\label{tab:few-shot-text-extraction-org-vs-cal} Results of few-shot learning experiments on bloom models with and without prompt calibration for text extraction tasks.}
\end{table*}

\begin{table*}[h!]
    \centering
    \begin{tabular}{lrrrrrrr}
        \hline
        \textbf{Models} & \textbf{Accuracy} & \textbf{en} & \textbf{es} & \textbf{de} & \textbf{fr} & \textbf{ja} & \textbf{zh}\\
        \hline
        gpt2-medium & $61.30$ & $66.7$ & $61.08$ & $65.26$ & $60.9$ & $58.04$ & $55.84$\\
        gpt2-xl & $55.25$ & $64.46$ & $56.58$ & $57.96$ & $56.62$ & $47.52$ & $48.38$\\
        bloom-560m & \boldmath$61.46$ & $65.38$ & $61.92$ & $63.46$ & $61.84$ & $58.28$ & $57.92$\\
        bloom-1b7 & $43.43$ & $43.32$ & $42.48$ & $47.96$ & $41.34$ & $40.6$ & $44.92$\\
        \hline
    \end{tabular}
    \caption{\label{tab:full_marc_task_scores} Full results of fine-tuning all the baselines on MARC}
\end{table*}

\begin{table}
    \centering
    \begin{tabular}{lp{3em}p{3em}p{3em}p{3em}p{3em}p{3em}}
        \hline
        \textbf{Models} & \textbf{toxicity} & \textbf{severe toxicity} & \textbf{obscene} & \textbf{threat} & \textbf{insult} & \textbf{identity attack}\\
        \hline
        gpt2 & $0.313$ & \boldmath$0.022$ & \boldmath$0.183$ & $0.019$ & $0.121$ & $0.037$\\
        gpt3 & \boldmath$0.331$ & $0.021$ & $0.178$ & $0.019$ & \boldmath$0.127$ & \boldmath$0.04$\\
        bloom-1b7 & $0.26$ & $0.016$ & $0.122$ & \boldmath$0.022$ & $0.101$ & $0.029$\\
        \hline
    \end{tabular}
    \caption{\label{tab:full_toxicitu_scores} Full results of toxicity analysis}
\end{table}

\section{Text Generation Samples}
\label{sec:appendix_text_generation}

\begin{table*}
    \centering
    \begin{tabular}{lp{34em}}
        \hline
        \textbf{Question} & \textbf{What is the purpose of endorsing a check?}\\
        \hline
        gpt2-medium & No, it's to make sure that we have a vote on which of the two of us will be supporting the senator. What is the purpose of the ad? There's a simple reason: we want senators who are in the minority to be\\ 
        gpt-neo & The purpose of endorsing a check is to provide the beneficiary with the funds that are required to pay the check. If the check is a cashier's check, the purpose is to provide the cashier with the funds necessary to pay the check. If the check is\\
        gpt2-xl & The purpose of endorsing a check is to give the checker the opportunity to check the validity of the check. If the checker receives a check that appears valid, then the check will be accepted. If the checker receives a check that appears invalid, then the check will\\
        bloom-350m & Just to keep your money safe. The only thing I can say is that the money you make goes to the families of the people who died and who have lost their lives because of the negligence of this government. If the government is just going to keep the money and\\
        bloom-1b3 & It is to be used against the debtor's account. The bank or other financial institution may also endorse a check, but it is the bank or other financial institution's decision.\\
        bloom-2b5 & To make sure the check is valid and to ensure that the person who wrote the check is the right person to receive it. You can read more about the purpose of endorsing a check here. When you sign a check, you must be able to see your signature clearly\\
        bloom-3b & What does the check tell the company that it is for the purpose of paying the client’s bill? If you are going to use a check as a payment method, you should make sure that the check is for the purpose of paying the bill and that the check is not\\
        bloom-7b1 & The purpose is to ensure that the payer will not get a check that is void or forged. If the payer has a legitimate reason to doubt the validity of the check, they can call the bank and ask for a duplicate check. The bank will send a duplicate check\\
        \hline
        \textbf{Question} & \textbf{Purpose of having good credit when you are well-off?}\\
        \hline
        gpt2-medium & The U.S. government has already had a very bad year for the credit rating agencies. The National Association of Realtors' (NASR) and Moody's have both been underrating the credit rating agencies, and the Credit Suisse\\ 
        gpt-neo & This is a very important question. I can't really answer it as I don't know the answer. But I will try to answer it. I am not saying that having good credit is bad. I am saying\\
        gpt2-xl & Is it the best way to buy a house? How can I get a loan for a house? How can I get a loan for a car? How can I get a loan for a car? How\\
        bloom-350m & The answer is no. But if you have bad credit, you still have to put in the time and effort to pay off your debts. You need to be able to pay off your debts in time to get the best possible results and minimize your\\
        bloom-1b3 & Well, it doesn’t have to be as bad as you think. If you are able to do so, you will be able to get more loans and make more money at the same time. The fact is that this is possible in order to make the application\\
        bloom-2b5 & Why do you need to have a good credit score? The answer is simple. You need a good credit score to get a mortgage, get a car loan, or get a credit card.\\
        bloom-3b & It is to be used against the debtor's account. The bank or other financial institution may also endorse a check, but it is the bank or other financial institution's decision.\\
        bloom-7b1 & It is more than a desire to have a nice car or a new house. Having good credit can help you get a job easily, get a loan, and even get a credit card. In other words, good credit is a tool for success. The\\
        \hline
    \end{tabular}
    \caption{Samples of generated text for questions in the FiQA dataset.}\label{tab:fiqa_text_1}
\end{table*}

\begin{table*}
    \centering
    \begin{tabular}{lp{34em}}
        \hline
        \textbf{Question} & \textbf{How many parties are there of the office holders whose one of the party is Janata Dal ?}\\
        \hline
        \textbf{Answer} & There are [47] other parties of the officeholders whose one party is Janata Dal.\\
        \hline
        gpt2-medium & The only party that has to be in the government is the party that is in power with the government and the party that is in power with the opposition is the party that is in power with the opposition.\\ 
        gpt-neo & It is the Janata Dal which has more than 3 lakh party workers in Punjab. I thought the party has more than 3 lakh party workers in Punjab. I think Dalit leaders are not so\\
        gpt2-xl & The answer is one. In fact, there are only one. So, there are only one chief minister in the entire country. The Congress is the party of the chief minister. But the Congress is not the party of the\\
        bloom-350m & And how many are there of the office holders whose one of the party is BJP? And how many are there of the office holders whose one of the party is JITC? And how many are there of the office holders\\
        bloom-1b3 & Janata Dal is a party which is ruling in the state. Janata Dal is not a party which is in the opposition. Janata Dal is a party which is ruling in the state. Janata Dal is a party\\
        bloom-2b5 & -- no response was generated --\\
        bloom-3b & -- no response was generated --\\
        bloom-7b1 & -- no response was generated --\\
        \hline
        \textbf{Question} & \textbf{What is the alma mater of the scientists known for String theory?}\\
        \hline
        \textbf{Answer} & The alma maters of the scientists known for the String theory are [answer].\\
        \hline
        gpt2-medium & The first alma mater of the scientific community was the California Institute for Space Studies in Pasadena on May 7, 1976. The name was chosen because the program was known for its efforts in the field of computerized biology, which was known for\\ 
        gpt-neo & I am not sure of the almahmata of the scientists. But I do know that there are many scientists who have made contributions in the field of science, so it is not too difficult to find out who the almah\\
        gpt2-xl & The string theory community is led by Prof. Lawrence Krauss, a theoretical physicist at the California Institute of Technology who has been at the forefront of the theory. Is there any truth to the theory that people can see the future?\\
        bloom-350m & What is the essence of the ideas and concepts that make up the concept of a fundamental theory? What are the most important factors that determine the validity of the fundamental theory? How do the fundamental theories emerge? And what is the nature of the fundamental theories?\\
        bloom-1b3 & The answer is probably not the University of Chicago, but rather the University of California, Berkeley. Born in Berkeley to a Jewish immigrant father and a Russian mother, he was the first child born to a Jewish family in the United States. While still\\
        bloom-2b5 & What is the history of the field, and how did it get started? What is the history of the field, and how did it get started? What is the history of the field, and how did it get started? What is the history of the\\
        bloom-3b & Is it MIT? Harvard? Berkeley? Princeton? Yale? Is it the University of California? The University of Southern California? The University of Iowa? The University of Chicago? University of Toronto? University of Virginia? University of California, Irvine? University\\
        bloom-7b1 & And what is the most prestigious university in the world? Which university has the largest number of Nobel Prizes in its history? What university has the most Nobel laureates? Which university has the most Nobel Prize winners? Which university has the most Nobel laure\\
        \hline
    \end{tabular}
    \caption{Samples of generated text for questions in the ParaQA dataset.}\label{tab:paraqa_text}
\end{table*}

\begin{table*}
    \centering
    \begin{tabular}{lp{34em}}
        \hline
        \textbf{Title} & \textbf{Stratasys and Eckhart sign three year agreement}\\
        \hline
        Original & Stratasys NASDAQ SSYS announces an exclusive three year collaboration with toolmaker Eckhart Eckhart s 3D printing lab in Michigan will use Stratasys Fortus 450mc E370 and Fortus 250mc printers The partnership will also promote the adoption of Industry 4 0 and the use of microsensors in 3D printed tools for smart diagnostics Stratasys shares are up 1 8 to 22 69 3D printing peers that could move on the news Proto Labs PRLB 1 6 3D Systems DDD 0 8 ExOne XONE Voxeljet VJET Now read\\
        \hline
        gpt2-medium & Stratasys and Eckhart sign three year agreement, but it's unclear if that agreement will include a deal for the remaining two years. If the deal is approved, Eckhart will retain his option to sign a deal before this season. He will remain free to sign with any team for the remainder of the season, but the terms of his option will be reduced by two years if the new deal is approved. The deal is expected to cost \$1.6 million per year, with a cap hit of \$2.9 million. Eckhart has a \$1.5 million signing bonus. The\\
        \hline
        gpt-neo & The companies have signed an agreement to operate a joint venture in Thailand in the 3 to 5 years. The companies are seeking a new strategic partner in Thailand to increase the scope of the business and expand the business. The aim of the Thai partnership is to work with Stratasys to develop and produce a new generation of flexible, modular and cost effective products for the automotive, aerospace and industrial sectors. The companies are seeking a new strategic partner in Thailand to work with Stratasys to develop and produce a new generation of flexible, modular and cost effective\\
        \hline
        gpt2-xl & Stratasys and Eckhart sign three year agreement to further develop the world's first 3D-printable prosthetic hand. ``The combination of the unique functionality of the 3D-printed prosthetic hand with the unique functionality of a hand that can be tailored to an individual's hand type and strength is the ultimate in design," said Eckhart. ``We are excited to bring the world's first 3D-printed prosthetic hand to market and believe that our technology will revolutionize the prosthetic industry." The 3D-printed prosthetic hand is designed to be an affordable, durable and user-friendly\\
        \hline
        bloom-350m & Stratasys and Eckhart sign three year agreement with the University of Sydney. The agreement includes the possibility of a joint project to develop an automated method for the analysis of samples from a large number of samples in a short time. The project will be funded by the Australian Research Council. The project will be conducted by the University of Sydney. The project is also funded by the NSW Department of Health and Social Care, which is also involved in the project. The project will be conducted by the University of Sydney. The project will be used to gain a better understanding of the effects of a wide variety of environmental contaminants on the growth of\\
        \hline
        bloom-1b3 & Stratasys and Eckhart sign three year agreement to improve collaboration and reduce cost The two companies have agreed to sign a three-year agreement to work together on a range of projects, including the development of the company’s new 3D printing machine, the X3D Rapid Prototyping System. The company will also work with Eckhart to develop a new product line of 3D printing solutions, including a new resin-based resin transfer moulding process and a new resin transfer moulding machine. The company is also working with Eckhart to develop a new resin transfer moulding system.\\
        \hline
    \end{tabular}
    \caption{Samples of generated articles.}\label{tab:article_text}
\end{table*}

\end{document}

%% file: sections/introduction.tex
In the last five years, the introduction of ELMo \cite{peters-etal-2018-deep}, GPT \cite{radford2018improving}, and BERT \cite{devlin-etal-2019-bert}, T5 \cite{raffel2020exploring} has led to pre-trained models being the basis for fine-tuning for downstream tasks. Primarily BERT-based models have been used as the starting point for various token and sequence classification tasks \cite{lee-etal-2017-end,munikar2019fine,hoang-etal-2019-aspect,NEURIPS2020_f23d125d,NEURIPS2021_33a854e2,su-etal-2022-roberta} and T5-based or GPT-style models have been used for generation, and zero-shot and few-shot learning \cite{yu2020few,gao2020making,goswamy-etal-2020-adapting,zhao2021calibrate,cho2021unifying,liu2021dexperts,Yang_Gan_Wang_Hu_Lu_Liu_Wang_2022,mi2022cins}, and dialogue\footnote{\url{https://openai.com/blog/chatgpt/}}.
 
BigScience Large Open-science Open-access Multilingual Language Model \cite{scao2022bloom}, or BLOOM is a 176 billion parameter open-source language model, trained on 46 natural languages, that for most of the training languages is the first language model with 100B+ parameters\footnote{\url{https://github.com/bigscience-workshop/bigscience/tree/master/train/tr11-176B-ml}}. 
Performance evaluation and benchmarks with BLOOM primarily focus on the 176B model variant \cite{scao2022bloom}; something that, even for evaluation, requires a large amount of computing resources and more than 3 TB of storage space. However, previous works have shown that the performance gained by using large model variants over the smaller ones is insignificant compared to the increase in computing resources and time required to train the large model \cite{Sanh2019DistilBERTAD,gpt-neox-20b}. 

In this paper, we investigate the performance of the smaller variants of BLOOM (350m, 560m, 1b1, 1b3, 1b7, 2b5, 3b, and 7b1), that have significantly lesser storage and memory requirements, on several popular NLP tasks. In doing this, we benchmark BLOOM against GPT-style generative and BERT-style models. This allows us to make pertinent observations about the impact of parameter size, prompt order, learning with zero or few instances, multilingual settings, and toxicity on performance across both model categories.

The closest work to our investigation is presented in \newcite{scao2022bloom}. However, while \newcite{scao2022bloom} focus on just zero- and few-shot learning experiments, we empirically study its performance across several diverse NLP tasks including every GLUE benchmark task \cite{wang2019glue}, question answering \cite{rajpurkar-etal-2016-squad}, zero-shot and few-shot learning using text classification and text extraction tasks \cite{zhao2021calibrate}, multilingual text classification \cite{conneau-etal-2018-xnli,marc_reviews}, named entity recognition \cite{wang-etal-2019-crossweigh}, and text generation \cite{kacupaj2021paraqa,10.1145/3184558.3192301}. We also compare the BLOOM model(s) with widely used BERT-based models and generative GPT style models on GLUE benchmark tasks and the multilingual XNLI task \cite{kacupaj2021paraqa}. The contributions of this work are as follows:

\begin{enumerate}

    \item \textbf{Text Classification} - We evaluate the smaller BLOOM model variants on text classification tasks to find that in - (i) prompt-based zero- and few-shot learning setup, the BLOOM variants result in an accuracy increase of 0.22\%-8.92\% over GPT models on at least one of zero- and few-shot experiments (ii) full fine-tuning-based setup, the smaller variant performs similarly to or better than the larger variant (\textit{bloom-1b7}) (Section \ref{sec:experiment_results}).

    \item \textbf{Question Answering} - We also evaluate the BLOOM model for the question-answering task using SQuAD datasets. The results show that the \textit{bloom-560m} model outperforms the \textit{gpt2-medium} model by $\sim$5 F1 points on SQuAD 1.1 and 7.95 F1 points on SQuAD 2.0 questions with an answer (Section \ref{subsubsec:squad_qa}).
    
    \item \textbf{Text Extraction} - In a prompt-based zero- and few-shot setting, we show that - (i) the \textit{bloom-1b7} model reports an increase in the accuracy by 0.5\%-11.82\% over GPT models on at least one of zero- and few-shot experiments, (ii) adding a single instance to the prompt helps BLOOM models more than GPT-style models, (iii) BLOOM is not as sensitive to the order of inputs in a prompt (as compared to GPT2 and GPT3), (iv) the performance of BLOOM almost always improves with the number of samples in few-shot learning; this is contrary to GPT where there is a lot more variability, and (v) the larger BLOOM variant performs better than the smaller variant (Section \ref{subsubsec:zero_few_shot})
    
    \item \textbf{Multilingual} - Zero-shot cross-lingual and multilingual fine-tuning experiments show that BLOOM is at par or worse than mono-lingual GPT-2 models (Section \ref{subsubsec:multi_lingual}).
    
    \item \textbf{Toxicity} - Finally, we also show that the text generated using BLOOM model variants using prompt-based input is 17\% less toxic compared to GPT models (Section \ref{subsubsec:toxicity}).
\end{enumerate}

%% file: sections/related_works.tex
The advent of Pre-trained language models revolutionized modern natural language processing.  Especially, the success of Masked Language Modeling (MLM) employed in BERT \cite{devlin-etal-2019-bert}, autoregressive training employed in GPT \cite{radford2019language}, and permutation-based training employed in XLNet \cite{yang2019xlnet} has led to the development of several derivatives over the last few years. The works of \citet{radford2019language} and \citet{brown2020language} have highlighted that pre-trained language models can perform practical tasks without additional training. Multiple works have demonstrated that the overall performance of a language model tends to improve with an increase in the number of parameters \cite{hestness2017deep,kaplan2020scaling,rae2021scaling,zeng2021pangu,wei2022emergent,hoffmann2022training,smith2022using,chowdhery2022palm}. LLMs have been shown to be able to perform new tasks based on a few demonstrations or natural language instructions. The approaches have also been extended to multilingual scenarios wherein the BERT model is pre-trained with Wikipedia of 100 different languages. \citet{pires2019multilingual} presented its superior zero-shot abilities for an NER task in Hinglish. \citet{llms_cannot_learn_longtailknowledge} show that large language models can recall information ubiquitous in the pre-training data but struggle to capture long-tail knowledge. In \citet{probing_toxiccontent}, authors employ logistic regression classifiers to probe English, French, and Arabic PTLMs and quantify the potentially harmful content.